\documentclass[conference]{IEEEtran}

\makeatletter

\def\ps@IEEEtitlepagestyle{%
  \def\@evenfoot{}%
}
\def\mycopyrightnotice{%
  {\footnotesize XXX-X-XXXX-XXXX-X/XX/\$XX.00~\copyright~20XX IEEE\hfill}
  \gdef\mycopyrightnotice{}
}

\usepackage{blindtext}
\usepackage{eso-pic}
\IEEEoverridecommandlockouts
\usepackage{cite}
\usepackage{amsmath,amssymb,amsfonts}
\usepackage{algorithmic}
\usepackage{graphicx}
\usepackage{textcomp}
\usepackage{xcolor}
\def\BibTeX{{\rm B\kern-.05em{\sc i\kern-.025em b}\kern-.08em
    T\kern-.1667em\lower.7ex\hbox{E}\kern-.125emX}}
    
\usepackage{eso-pic}
\newcommand\AtPageUpperMyright[1]{\AtPageUpperLeft{%
 \put(\LenToUnit{0.17\paperwidth},\LenToUnit{-2cm}){%
     \parbox{0.9\textwidth}{\raggedleft\fontsize{8}{11}\selectfont #1}}%
 }}%
\newcommand{\conf}[1]{%
\AddToShipoutPictureBG*{%
\AtPageUpperMyright{#1}
}
}

\begin{document}
\title{\vspace*{1cm} Multi-agent Auto-Bidding with Latent Graph Diffusion Models.
}

\author{\IEEEauthorblockN{Dom Huh}
\IEEEauthorblockA{\textit{Department of Computer Science} \\
\textit{University of California, Davis}\\
Davis, California, United States \\
dhuh@ucdavis.edu}
\and
\IEEEauthorblockN{Prasant Mohapatra}
\IEEEauthorblockA{\textit{Department of Computer Science} \\
\textit{University of California, Davis}\\
Davis, California, United States \\
pmohapatra@ucdavis.edu}}

\maketitle

\conf{\textit{Preprint}}

\begin{abstract}
This paper presents a novel diffusion-based auto-bidding framework that uses graph representations to model large-scale auction environments. In such environments, agents must optimize bidding strategies dynamically, balancing key performance indicators (KPIs) while navigating competitive, uncertain, and sparse conditions. To address these challenges, we introduce an approach that combines learnable graph embeddings with a planning-based latent diffusion model (LDM). This model captures the intricate relationships between impression opportunities and multi-agent interactions, enabling more accurate predictions of auto-bidding outcomes. Through reward-alignment techniques, the LDM’s posterior is fine-tuned to maximize KPI performance under predefined constraints. Our experiments, conducted in both real-world and synthetic auction environments, show significant improvements in the accuracy of auction outcome forecasts through learnable graph-based embeddings and in bidding performance across a range of common KPIs.
\end{abstract}


\begin{IEEEkeywords}
auction, auto-bidding, graph, diffusion model
\end{IEEEkeywords}

\section{Introduction}
Optimal auto-bidding in large-scale auctions remains an open research challenge with significant real-world implications for online markets, particularly in domains like digital advertising. In these dynamic auction environments, auto-bidding agents compete by placing bids on high-volume, stochastic streams of impression opportunities (IOs). The objective is to maximize the value of the bids while adhering to multiple key performance indicators (KPI) constraints, such as budget adherence, return on investment (ROI), conversion rates (CVR), social welfare, and cost-per-acquisition (CPA) \cite{aggarwal2024autobiddingauctionsonlineadvertising, kitts2017ad}. The inherent uncertainty and competition in these auctions make the optimization task particularly difficult \cite{huh2024multiagentreinforcementlearningcomprehensive}, as agents must not only understand these interactions but also constantly adapt their strategies to account for changing conditions.

Recent advancements have shown the potential of generative diffusion models \cite{sohl2015deep} to simulate auction trajectories and plan bids, with a focus on maximizing KPI adherence \cite{guo2024generative}. These methods have demonstrated success in capturing complex, high-dimensional patterns in bidding strategies. However, current approaches primarily rely on heuristic feature representations that fail to capture the full extent of the interdependencies between IOs and the dynamic, multi-agent nature of auction environments. These limitations often hinder the ability to model the intricate relationships between agents, auction parameters, and impression opportunities, leading to suboptimal bidding strategies.

\begin{figure}
    \centering
    \includegraphics[width=\linewidth]{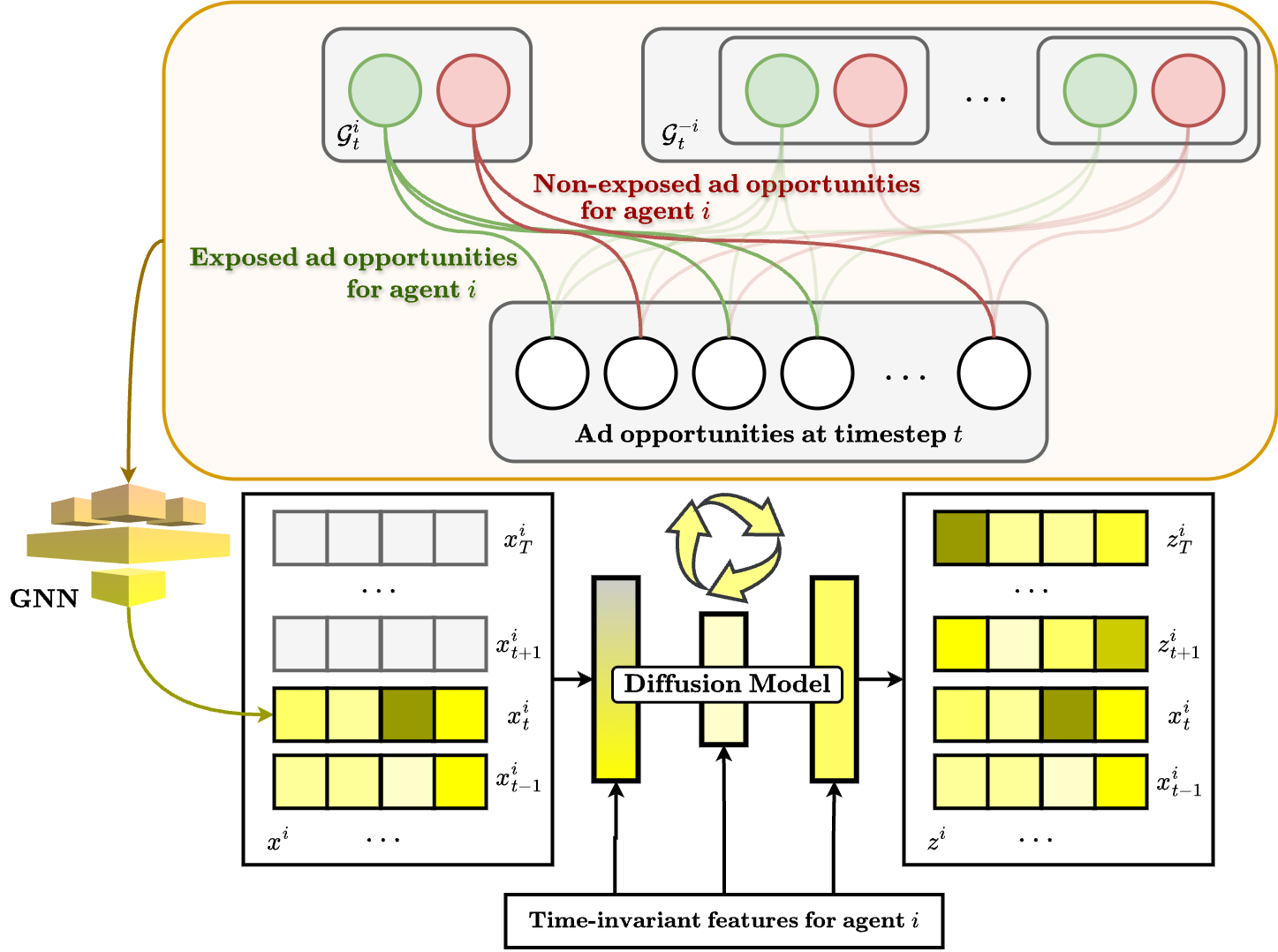}
    \caption{An illustration of the LGD-AB framework. At every time-step, a graph-based embedding is computed for agent $i$ using a bipartite graph of agent and IO nodes. The nodes of a sub-graph $\mathcal{G}_t^i$ consist of two virtual nodes, connecting to IOs exposed and not exposed to agent $i$ respectively. Sub-graphs for other agents are similarly formed and connected accordingly. The graph is processed using a GNN to generate an embedding vector $x^i_t$. The diffusion model then learns a posterior of the temporal sequence of embedding vectors to forecast auction dynamics.}
    \label{fig:enter-label}
\end{figure}

To address these challenges, we propose a novel framework that incorporates graph-based representations to model auction environments and a latent diffusion model (LDM) to plan and optimize bidding strategies. Our framework, Latent Graph Diffusion Model for Auto-Bidding (LGD-AB), goes beyond traditional heuristics by using learnable graph embeddings to represent the complex interrelationships between IOs, agents, and auction outcomes by modeling IOs on a more granular level to retain their individual attributes as well as encoding their connections to other IOs and agents. This enables more expressive computations and a deeper understanding of the auction dynamics. Furthermore, we use a reward alignment technique \cite{huh2025maximizediffusionstudyreward} to optimize multiple KPI metrics simultaneously \cite{he2021unified}, forming a multi-objective optimization framework.

The contributions of this paper are twofold: first, we introduce a graph-based approach to auto-bidding that captures the complex, multi-agent interactions in auction environments; second, we propose the use of latent diffusion models for generating optimized bidding trajectories. Through extensive empirical evaluations on both synthetic \cite{deng2024non} and real-world auction datasets \cite{su2025auctionnet}, we demonstrate that our method significantly improves auto-bidding performance across several KPI metrics, as well as forecasting accuracy for auction outcomes.

\begin{figure}
    \centering
    \includegraphics[width=\linewidth]{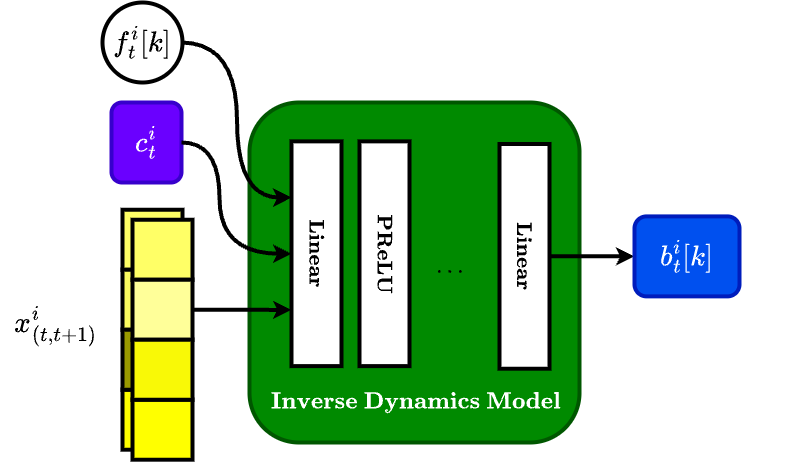}
    \caption{Inverse Dynamics Model for Bid Computation.}
    \label{fig:idm}
\end{figure}

\section{Latent Graph Diffusion Model for Auto-Bidding} 
Our proposed framework, LGD-AB, comprises two core components: (1) a graph-based embedding module for encoding auction dynamics and (2) a multi-agent latent diffusion model for auction forecasting and strategic bid optimization.

\subsection{Graph Embedding For Ad Auctions}
The objective of the graph embedding is to represent the IOs presented to each agent in an expressive and comprehensive manner. In large-scale auctions, the number of IOs exposed to each agent can range drastically, from $0$ to $10$s of thousands at every time-step. To process data with such variability yet interdependency, we represent these IOs as nodes of a graph and process their features using graph convolutions with an attention-based graph neural network (GNN) \cite{brody2022attentivegraphattentionnetworks}. Rather than using a relational GNN, we instantiate two virtual nodes for each agent to connect exposed and non-exposed IOs with undirected links separately. By using bidirectional links between IOs and agent nodes, this design choice enables information propagation between agents with multiple hops. The additional links dedicated to non-exposed IOs aim to maintain context during sparse time periods. For scalability and computational limitations for larger scale simulations (e.g., dense graphs with $>1$ million edges), we perform random neighbor sampling on the non-exposed links to avoid prohibitively large computations, as opposed to a hierarchical graph building approach \cite{huh2020hierarchicalbigraphneuralnetwork}. To compute the agent embedding, we aggregate the node embeddings of the two virtual nodes with a summation operation.

We optimize the latent graph embedding $x^i_t$ of agent $i$ at time-step $t$ generated by the GNN to contain predictive information regarding the bids $b_t^i[k]$ made for each IO $k$. Hence, we introduce a separate inverse dynamics model (IDM) to support the optimization of the following objective:
\begin{equation}\label{eq: loss}
    \mathcal{L}_{\text{graph}}(\theta) = \mathbb{E}_{(t,k)}[\log \text{P}(b_t^i[k] | \theta, x_t^i,  x_{t+1}^i, f_t^i[k], c_t^i)]
\end{equation}
where $k$ is the IO identification index and $f_t^i[k]$ are the node embedding of the corresponding IO. To ensure the feasibility of this optimization, we append a context vector $c_t^i$ that provides more information regarding the agent agnostic to the individual IO, as shown in Figure \ref{fig:idm}. We optimize the objective over the parameters $\theta$ of both the GNN and the IDM. We view this optimization to (1) instill bid-policy information into the embedding and (2) constrain the graph embedding to reflect auction dynamics.

\begin{figure}
    \centering
    \includegraphics[width=\linewidth]{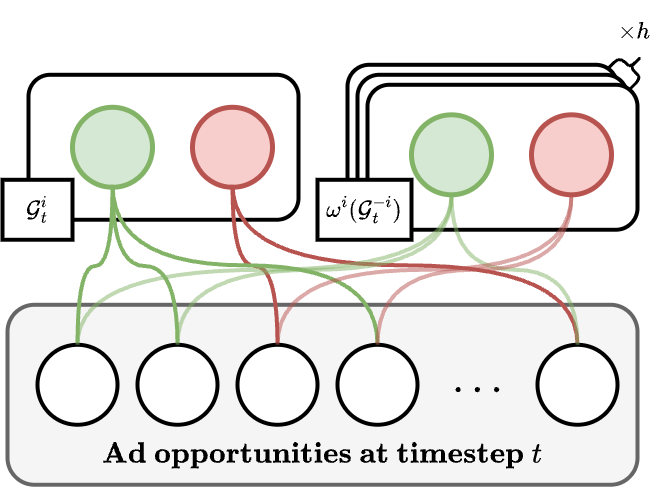}
    \caption{Agent $i$'s Approximation of Other Agents' Graph $\omega^i(\mathcal{G}_t^{-i})$ by creating $h$ subgraphs used in student model $\mathcal{S}$.}
    \label{fig:belief graph}
\end{figure}

\paragraph{Incomplete Information} The advertiser's visibility of other agents' IOs and bids is contingent on the auction design, e.g., sealed and closed envelope auctions, which could prevent a complete and accurate construction of the subgraphs of other agents. In such cases, we introduce a belief graph $\omega^i(\mathcal{G}_t^{-i})$, which approximates the other agents' sub-graph by creating $h$ agent subgraphs, as shown in Figure \ref{fig:belief graph}. For our experiments, we set $h = 4$. For simplicity, the nodes are randomly linked, where IO nodes are partitioned between the exposed and non-exposed nodes. To train the embedding, we apply knowledge distillation (KD) \cite{Gou_2021}. Therefore, we train a centralized GNN $\mathcal{T}$ using Equation \ref{eq: loss} with full visibility of all agents' sub-graphs. We then train a student model $\mathcal{S}$ using a response-based KD loss to mimic the outputs of $\mathcal{T}$. 

\paragraph{Multi-Agent Latent State Optimization} As defined so far, the embedding vector does not account for the temporal nature of the auction dynamics. Using a recurrent model representation or stacking multiple time steps are common approaches to handle such discrepancies. However, such methods are not scalable, nor can they handle open environments (e.g., advertisers can enter and leave the auction) trivially. Hence, we extend representation learning techniques, namely MA-SPL \cite{huh2024representationlearningefficientdeep}, to embed multi-agent and temporal dynamics information within the latent graph embedding $\{x^i_0, x_1^i, \dots, x_T^i\}_{i\in A}$, where $A$ is the set of all agents. 

\begin{figure}
    \centering
    \includegraphics[width=\linewidth]{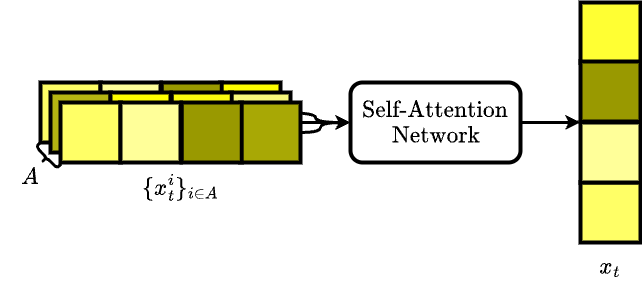}
    \caption{Computing Joint Embedding Vector using Self-Attention Network.}
    \label{fig:eq}
\end{figure}

\paragraph{Equilibrium Computation}
To promote a more cohesive multi-agent bidding strategy, we compute a joint embedding space by merging all agents' graph embeddings. We refer to this design choice as equilibrium computation (EC), as the computation of all agents (i.e., bid computation and planning) is rooted in a shared latent state. Hence, we introduce an aggregation operation $\mathcal{A}: \{x^i_t\}_{i\in A} \mapsto x_t$ using a self-attention network, as shown in Figure \ref{fig:eq}. The self-attention network follows the Transformer encoder with a global mean pooling. This latent embedding $x_t$ is now shared between all agents when computing the generative objectives and for individual bid formulation.

\subsection{Latent Diffusion Models}
The latent diffusion model (LDM) processes a temporal sequence of graph embeddings $\{x_t\}_{t\in[\tau,\tau+T]}$, where $T$ is the length of the context window, and forms the foundation to devise a planning-based auto-bidding solution. In this work, we adopt the Decision Diffuser framework \cite{ajay2023is}, denoising state trajectories and offloading the action generation to the IDM. Unlike recent works using transformer-based architectures \cite{peebles2023scalable} and score-based prediction \cite{song2020score}, we found greater empirical success using a more traditional 1D temporal convolution network and denoising with $\epsilon$-prediction. As shown in Figure \ref{fig:enter-label}, time-invariant features of the agent, such as the agent's category and total budget, are processed as conditional inputs. We optimize the diffusion objective using in-painting masks \cite{lugmayr2022repaint}.

\paragraph{Reward Alignment} To ensure that the LDM's posterior aligns with the KPI metrics and auto-bidding objective, we follow the procedure set in \cite{huh2025maximizediffusionstudyreward}, utilizing an iterative sequence of constrained alignment optimization of reinforcement learning, direct preference optimization, and supervised fine-tuning with rejection sampling. We train a value function using IQL \cite{kostrikov2021offline} to predict not only the value, i.e., the number of total conversations, but also other important KPI metrics, such as CPA, ROI, win rate, budget adherence, and social welfare \cite{deng2024non}.
\begin{itemize}
    \item Cost per acquisition (CPA) - CPA is the ratio between the total cost incurred and the earned value earned over all IOs, measuring the general performance of the auto-bidding strategy.
    \item Return on Investment (ROI) - ROI measures the profitability of the strategy, computed as the ratio between the net profit and the total cost.
    \item Win Rate - Win rate measures the effectiveness of a bidding strategy or decision-making process in securing wins and is computed as the ratio between the number of winning bids and the number of total bids made.
    \item Budget Adherence - Budget adherence measures how well the spending aligns with the allocated budget. This metric ensures that resources are used efficiently and within the expected limits and is computed as the exceedance rate of the budget spending.
    \item Social Welfare - Social welfare represents the aggregate value or utility generated by the system, measured as the total earned value over all advertisers.
\end{itemize}
With this value function, we can also strictly enforce KPI constraints, e.g., budget adherence, through rejection sampling.

\paragraph{Fine-Tuning} Training LGD-AB occurred in two stages. We first trained the graph embedding module and then the LDM separately. We then train both modules together, fine-tuning the GNN at a lowered learning rate. We found that having separate training stages was much more stable, and the additional stage of joint training benefits the performance of both modules. For partial observable settings, we train the teacher model using the described approach and then train the student model afterward.

\section{Results and Ablations Studies}

\begin{table*}
\caption{Comparing performances between DiffBid and LGD-AB  ($\mathcal{S}$, EC, and post-alignment) with mean and $\pm 1$ std over $64$ random seeds.}
    \centering
    \begin{tabular}{|c|c|c|c|c|}
     \hline
& \multicolumn{2}{c|}{AuctionNet} & \multicolumn{2}{c|}{Synthetic Auction}   \\ \cline{2-5} 
    & DiffBid & LGD-AB & DiffBid & LGD-AB  \\ \hline
\multicolumn{1}{|c|}{Return} & $353.19 \pm 0.42$ & $489.04 \pm 1.3$  & $982.59 \pm 10.61$ & $2123.10 \pm 15.32$ \\ \hline
\multicolumn{1}{|c|}{CPA} & $1.041 \pm 0.028$ & $0.992 \pm 0.009$ & $0.842 \pm 0.014$ & $0.592 \pm 0.010$ \\ \hline
\multicolumn{1}{|c|}{ROI} & $0.012 \pm 0.001$ &  $0.030 \pm 0.001$ & $0.026 \pm 0.001$ & $0.055 \pm 0.001$ \\ \hline
\multicolumn{1}{|c|}{Win Rate} & $0.071 \pm 0.002$ & $0.152 \pm 0.008$ &  $0.089 \pm 0.001$ & $0.203 \pm 0.003$ \\ \hline
\multicolumn{1}{|c|}{Budget Adherence} & $0.995 \pm 0.002$ & $0.995 \pm 0.002$ & $0.998 \pm 0.001$ & $1.000 \pm 0.000$ \\ \hline
\multicolumn{1}{|c|}{Social Welfare} & $3.63 \pm 0.81$ & $4.09 \pm 0.07$ & $50.43 \pm 1.42$ & $69.20\pm 0.56$ \\ \hline
\end{tabular}
\label{tab:kpi}
\end{table*}

\subsection{Experiment Setup}
We evaluate LGD-AB on two auto-bidding simulations: AuctionNet \cite{su2025auctionnet} and synthetic auction \cite{deng2024non}.

\paragraph{AuctionNet} AuctionNet \cite{su2025auctionnet} contains over $500$ million in real-world records of online advertising instances, with $21$ separate bidding periods of $48$ time-steps and $48$ different advertisers. We use an auto-bidding simulator, following prior works \cite{chiappa2024auto}, which reenacts auction dynamics using the dataset by having the auto-bidding agent impersonate one of the $48$ advertisers.

\paragraph{Synthetic Auction} Synthetic auction, on the other hand, generates artificial auction data that mimics real ad auctions by capturing their underlying properties through hierarchical feature generation and specific probabilistic distributions. We extend a prior work by Deng et al. \cite{deng2024non} with configurable parameters of IOs to introduce longer and variable ad life cycles, multi-slot winning IOs, as well as a budget constraint on each agent. We collect an offline RL dataset of $100$ thousand auction simulations with a varying number of advertisers, allocation rule (e.g., FPA, GSP, VCG), bidding period durations, and IO's properties mentioned above. Within this dataset, advertisers randomly select a uniform bid-scaling strategy or a non-uniform bid-scaling algorithm with pre-computed parameters \cite{deng2024non}. 

\subsection{Empirical Evaluation}

\begin{table}
    \centering
        \caption{Auction Forecasting Results of LGD-AB Design Choices.}
    \begin{tabular}{|c|c|c|}
     \hline
& \multicolumn{2}{c|}{$\mathcal{L}_{fc}(\theta)$}   \\ \cline{2-3} 
    & AuctionNet & Synthetic Auction    \\ \hline
\multicolumn{1}{|c|}{$\mathcal{T}$} & $3.43$ & $2.59$ \\ \hline
\multicolumn{1}{|c|}{$\mathcal{S}$} & $2.64$ & $2.02$ \\ \hline \hline
\multicolumn{1}{|c|}{No EC}        & $3.37$ & $2.13$ \\ \hline
\multicolumn{1}{|c|}{EC}        & $3.32$ & $2.62$ \\ \hline \hline
\multicolumn{1}{|c|}{No Alignment} & $3.25$ & $2.09$ \\ \hline
\multicolumn{1}{|c|}{Alignment}  & $3.02$ & $2.04$ \\ \hline
    \end{tabular}
    \label{tab:Forecasting}
\end{table}

\begin{figure*}
    \centering
    \includegraphics[width=\linewidth]{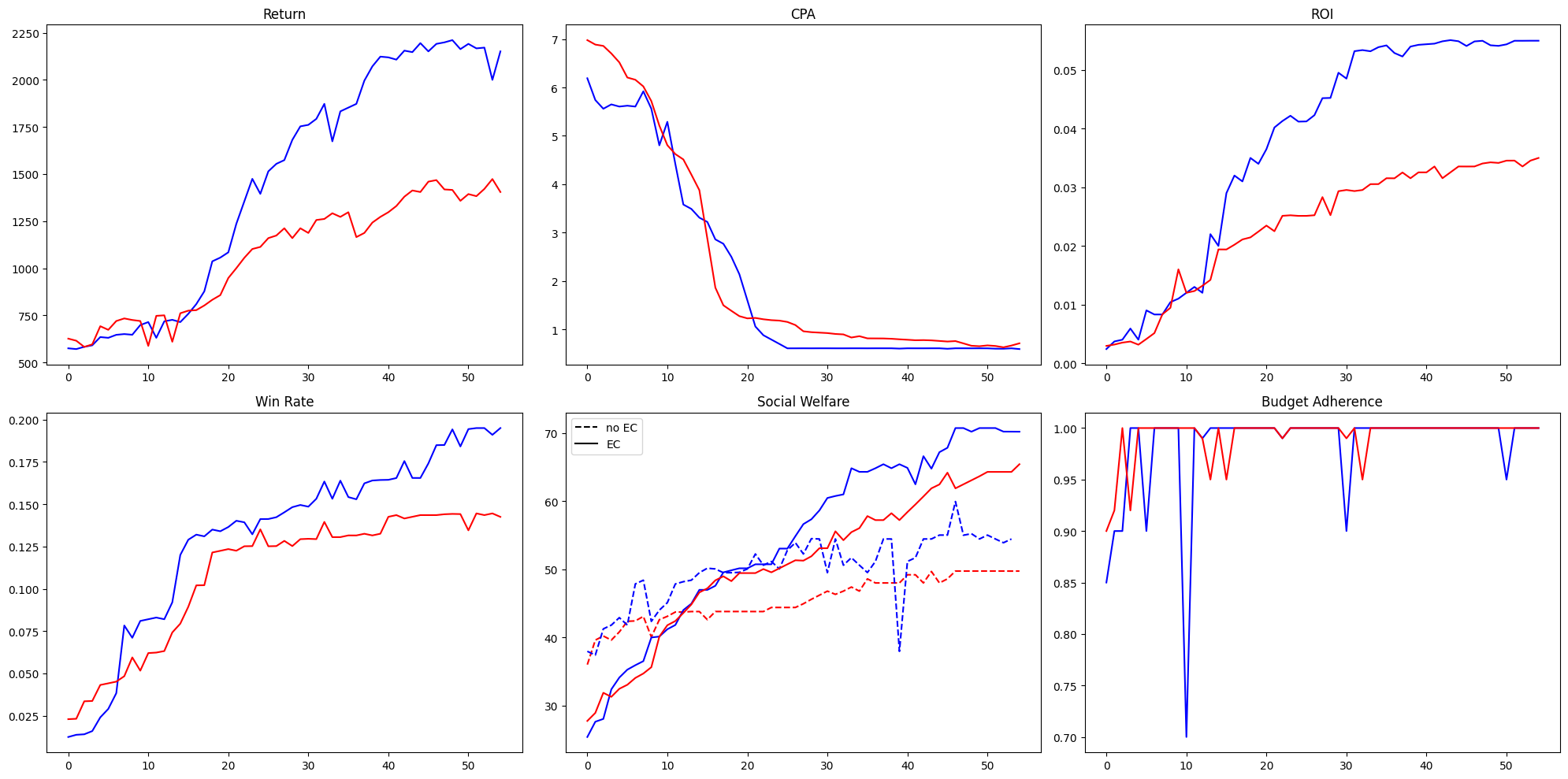}
    \caption{KPI Alignment Learning Curve on Synthetic Auction with multi-KPI alignment (shown in red) compared to optimizing each KPI criterion independently (shown in blue). For social welfare, we compare the training between EC and non-EC variants, as shown in the dotted lines.}
    \label{fig:kpi}
\end{figure*}

\paragraph{Auction Forecasting} We measure the forecasting capabilities of our LGD-AB framework by comparing the ground truth outcomes $x_{t'>t}$ with the predictions of the LDM $z_{t'>t}$ given a masked input $x_{t'\leq t}$. To quantify this measure $F(\theta)$, we compute the log-likelihood of the LDM's posterior to the path towards ground truth outcomes.
\begin{equation}
    \mathcal{L}_{fc}(\theta) = \mathbb{E}_{x, t}[\log \text{P}(x_{t'>t} | \theta, x_{t'\leq t})]
\end{equation}
In Table \ref{tab:Forecasting}, our evaluation was performed using our online simulators, where to compute $F(\theta)$, we use a testing dataset that was not used during training. We found that even with KD only, the student model retains, on average, $77.48\%$ of the teacher model's forecasting accuracy despite the limitations placed by the partial observability. EC also improved forecasting capability in Synthetic Auction; however, in AuctionNet, it failed to replicate this improvement. We attribute this failure to the limited number of joint data samples (i.e., only $21$ bidding periods in total). To assist with this limitation and mitigate the efforts from data scarcity, we curated and trained on synthetic data generated by the LDM, utilizing a quality-based rejection sampling. By training on this synthetic dataset, we found a more fair representation of its forecasting accuracy, achieving a score of $0.21$ to $3.32$. Lastly, the KPI alignment reduced average forecasting accuracy by $9.47\%$, which is expected, as such fine-tuning may lower the likelihood of lower-value auto-bidding outcomes.

\paragraph{KPI and Performance} We compare our LGD-AB framework to another diffusion-based auto-bidding solution, DiffBid \cite{guo2024generative}, across six common KPI metrics. We highlight that the main differences of the DiffBid framework to our LGD-AB are its use of a heuristic feature space, bid-scaling, and a conditional return input. In the original work of DiffBid, the authors have made comparisons to other competitive generative models (i.e., Decision Transformer) and RL baselines (i.e., BC, CQL, and IQL) to DiffBid, where DiffBid has demonstrated to outperform these other methods on similar auto-bidding simulations.

From Table \ref{tab:kpi}, we observe improvements across all KPIs using our LGD-AB framework from DiffBid. The results demonstrate a significant increase in the number of IOs won, shown by an average $\times1.29$ improvement in return from our competitive baseline. This was achieved while significantly maximizing the win rate, lowering CPA, and maintaining budget adherence. Additionally, our proposed LGD-AB has demonstrated better social welfare, indicating an overall higher collective utility to the auction. In both our baseline and LGD-AB solution, budget adherence was not an issue and remained comparable, maintaining near-perfect budget adherence over all trials. We note that budget adherence was applied as a constraint via rejection sampling with a learned function with fair empirical risk and was not optimized using the alignment methods.

In Figure \ref{fig:kpi}, we find that optimizing over multiple KPI (i.e., return, CPA, ROI, win rate, and social welfare) at the same time provides meaningful improvements within the individual performances of each metric. For some metrics, the difference is more significant, such as return and ROI, which are quite important measures in real-world applications.

Lastly, while EC did not show significant improvements over all KPI metrics, we found its utility in enabling much healthier alignment of social welfare.

\begin{table}
    \centering
    \caption{Comparing Bid Computation Accuracy of DiffBid and LGD-AB.}
    \begin{tabular}{|c|c|c|}
     \hline
& \multicolumn{2}{c|}{Average $\ell_2$ Norm per Agent}   \\ \cline{2-3} 
    & AuctionNet & Synthetic Auction    \\ \hline
\multicolumn{1}{|c|}{DiffBid} & $19.78$ & $82.10$ \\ \hline
\multicolumn{1}{|c|}{LGD-AB} & $13.41$ & $37.82$ \\ \hline
    \end{tabular}
    \label{tab:bid_accuracy}
\end{table}

\paragraph{Accuracy of Bid Computation} We compare the bid prediction accuracy of DiffBid and LGD-AB in Table \ref{tab:bid_accuracy}. The results show a noticeably lower $\ell_2$ error with our LGD-AB's $\ell_2$, alluding to being able to extract bids more accurately from our graph embeddings than the pre-constructed feature heuristic and limitations of the bid-scaling auto-bidding approaches. Notably, in the Synthetic Auction simulation dataset, where portions of the auction data are generated from non-uniform bid-scaling strategies, we notice a wider margin of difference in accuracy in favor of our LGD-AB framework.

\paragraph{Limitations and Future Directions} We observe some key issues and limitations while developing our proposed solution, mainly pertaining to the scale of data and efficiency of our generative modeling algorithm. With AuctionNet, while there exists a large set of individualized auction records to train on, there was an insufficient number of collective auction records that span full bidding periods to properly train a generative model to capture the underlying auction dynamics accurately. We addressed this limitation with synthetic data generation and hyperparameter tuning on the diffusion model. However, this limited performance can be improved with advancements in either training generative models in low data regimes (e.g., few-shot learning) or greater data availability. Another issue concerns the computational limitation from scaling the graph representation w.r.t. the number of IOs and the number of advertisers at every time-step, which was handled using neighbor sampling, which limits expressivity and completeness in the embedding computation. In future works, a form of dynamic graph sparsification or hierarchical graph representations could mitigate this. Lastly, a key 

\section{Conclusion}
In this work, we propose a latent diffusion planning framework using graph-based embedding for auto-bidding applications. We demonstrate the utility of a graph-based representation, which enabled the LDM to capture the underlying dynamics distribution of a multi-agent auction process. By aligning various KPIs with the trained LDM, we devised an effective bidding strategy that is capable of generating bids that maximize various KPI metrics.

\bibliography{citations}
\bibliographystyle{ieeetr}

\end{document}